\documentclass[nohyperref]{article}

\usepackage{microtype}
\usepackage{graphicx}
\usepackage{booktabs} 
\usepackage{tikz}
\usepackage{pgfplots}
\pgfplotsset{compat=1.17}
\usepackage{subcaption}
\usepackage{hyperref}

\usepackage[accepted]{icml2022}
\usepackage{amsmath}
\usepackage{amssymb}
\usepackage{mathtools}
\usepackage{amsthm}
\usepackage[capitalize,noabbrev]{cleveref}

\usepackage{float}

\theoremstyle{plain}

\theoremstyle{definition}

\theoremstyle{remark}

\usepackage[textsize=tiny]{todonotes}

\icmltitlerunning{Branchformer: Parallel MLP-Attention Architectures for Speech Recognition and Understanding}

\begin{document}

\twocolumn[
\icmltitle{Branchformer: Parallel MLP-Attention Architectures to Capture \\
Local and Global Context for Speech Recognition and Understanding}



\icmlsetsymbol{equal}{*}

\begin{icmlauthorlist}
\icmlauthor{Yifan Peng}{ece}
\icmlauthor{Siddharth Dalmia}{lti}
\icmlauthor{Ian Lane}{ece}
\icmlauthor{Shinji Watanabe}{lti}
\end{icmlauthorlist}

\icmlaffiliation{ece}{Department of Electrical and Computer Engineering, Carnegie Mellon University, Pittsburgh, PA 15213, USA}
\icmlaffiliation{lti}{Language Technologies Institute, Carnegie Mellon University, Pittsburgh, PA 15213, USA}

\icmlcorrespondingauthor{Yifan Peng}{yifanpen@andrew.cmu.edu}
\icmlcorrespondingauthor{Siddharth Dalmia}{sdalmia@cs.cmu.edu}
\icmlcorrespondingauthor{Shinji Watanabe}{swatanab@andrew.cmu.edu}

\icmlkeywords{Machine Learning, ICML}

\vskip 0.3in
]



\printAffiliationsAndNotice{}  

\begin{abstract}
Conformer has proven to be effective in many speech processing tasks. It combines the benefits of extracting local dependencies using convolutions and global dependencies using self-attention.
Inspired by this, we propose a more flexible, interpretable and customizable encoder alternative, \textit{Branchformer}, with parallel branches for modeling various ranged dependencies in end-to-end speech processing. In each encoder layer, one branch employs self-attention or its variant to capture long-range dependencies, while the other branch utilizes an MLP module with convolutional gating (cgMLP) to extract local relationships. We conduct experiments on several speech recognition and spoken language understanding benchmarks. Results show that our model outperforms both Transformer and cgMLP. It also matches with or outperforms state-of-the-art results achieved by Conformer.
Furthermore, we show various strategies to reduce computation thanks to the two-branch architecture, including the ability to have variable inference complexity in a single trained model. 
The weights learned for merging branches indicate how local and global dependencies are utilized in different layers, which benefits model designing.
\footnote{Our code and models are released as part of the ESPnet toolkit: \href{https://github.com/espnet/espnet}{https://github.com/espnet/espnet}.}

\end{abstract}

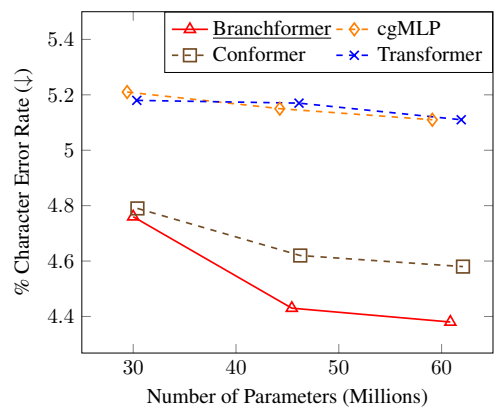
\begin{figure}[t]
\centering
\resizebox {0.8\linewidth} {!} {
\begin{tikzpicture}
	\begin{axis}[
		xlabel=Number of Parameters (Millions),
		ylabel=\% Character Error Rate ($\downarrow$),
		xtick={20,30,...,100},
		xmin=25,
        xmax=65,
        ymax=5.5,
	    every axis plot/.append style={thick},
        legend cell align={left},
        legend columns=2,
        legend style={at={(1,1)},anchor=north east}
		]
	\addplot[color=red, mark=triangle, mark options={scale=1.5}] coordinates {
	    (30.00,  4.76)
	    (45.43,  4.43)
	    (60.85,  4.38)
	}; 
	\addplot[color=orange, mark=diamond, dashed, mark options={scale=1.5, solid}] coordinates {
		(29.41,  5.21)
	    (44.26,  5.15)
	    (59.10,  5.11)
	};
	\addplot[color=brown!60!black,mark=square, dashed, mark options={scale=1.5, solid}] coordinates {
		(30.41,  4.79)
	    (46.25,  4.62)
	    (62.08,  4.58)
	};
	\addplot[color=blue,mark=x, dashed, mark options={scale=1.5, solid}] coordinates {
		(30.35,  5.18)
	    (46.13,  5.17)
	    (61.91,  5.11)
	};
    \legend{\underline{Branchformer},cgMLP,Conformer,Transformer}
	\end{axis}
\end{tikzpicture}
}
  \caption{Character Error Rate (\%) vs. Model Size. Our Branchformer outperforms previously proposed Conformer, cgMLP and Transformer at all scales on the benchmark Aishell ASR task.}
  \label{aishell-scaling}
  \vskip -0.1in
\end{figure}

\begin{figure*}[ht]
\vskip -0.1in
\centering
\hfill
\begin{subfigure}[t]{0.33\textwidth}
\includegraphics[width=0.65\textwidth]{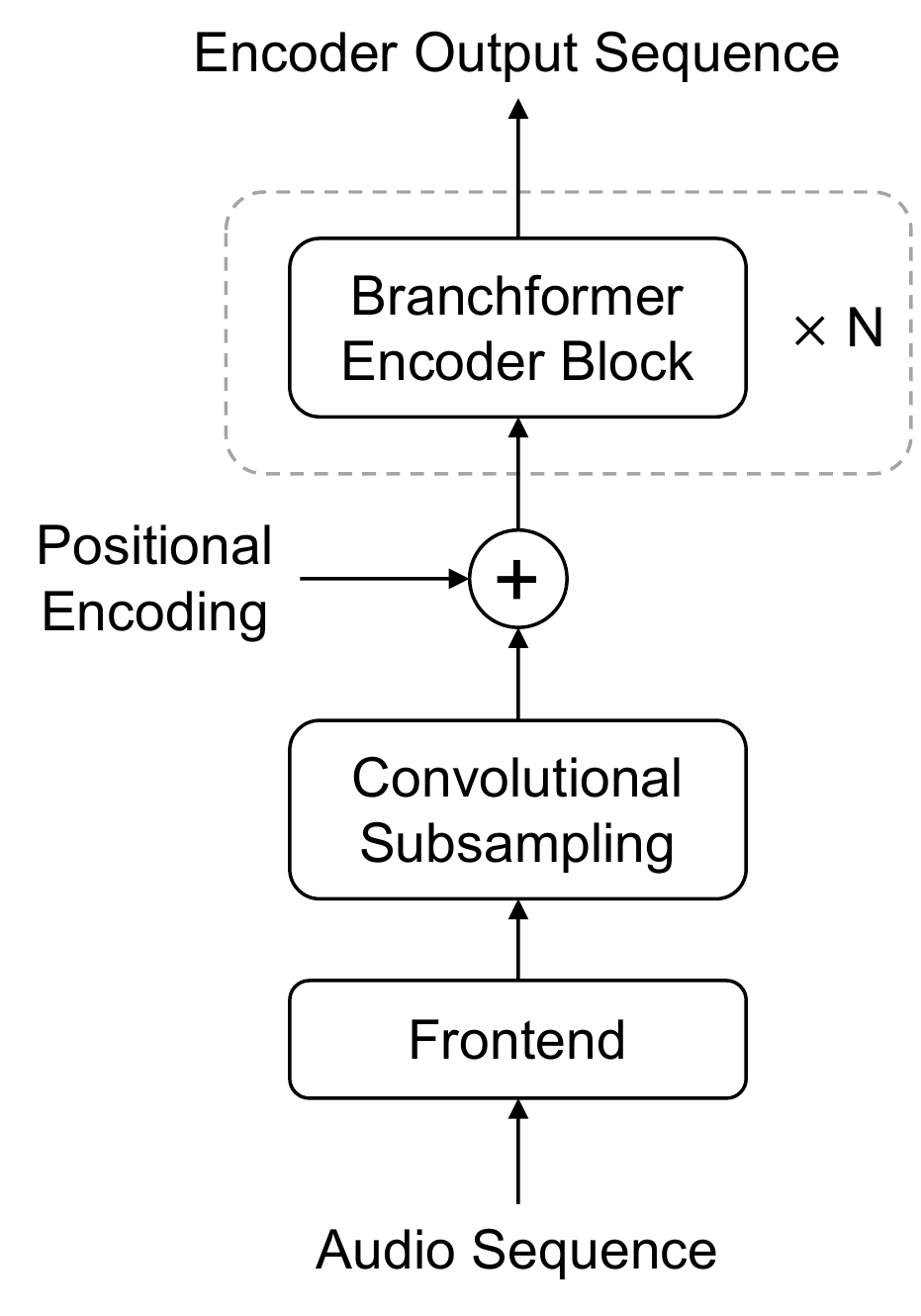}
\caption{Overall architecture of the encoder. A stack of identical Branchformer blocks are used to capture local and global dependencies.}
\label{encoder-arch}
\end{subfigure}
\hfill
\begin{subfigure}[t]{0.62\textwidth}
\includegraphics[width=0.9\textwidth]{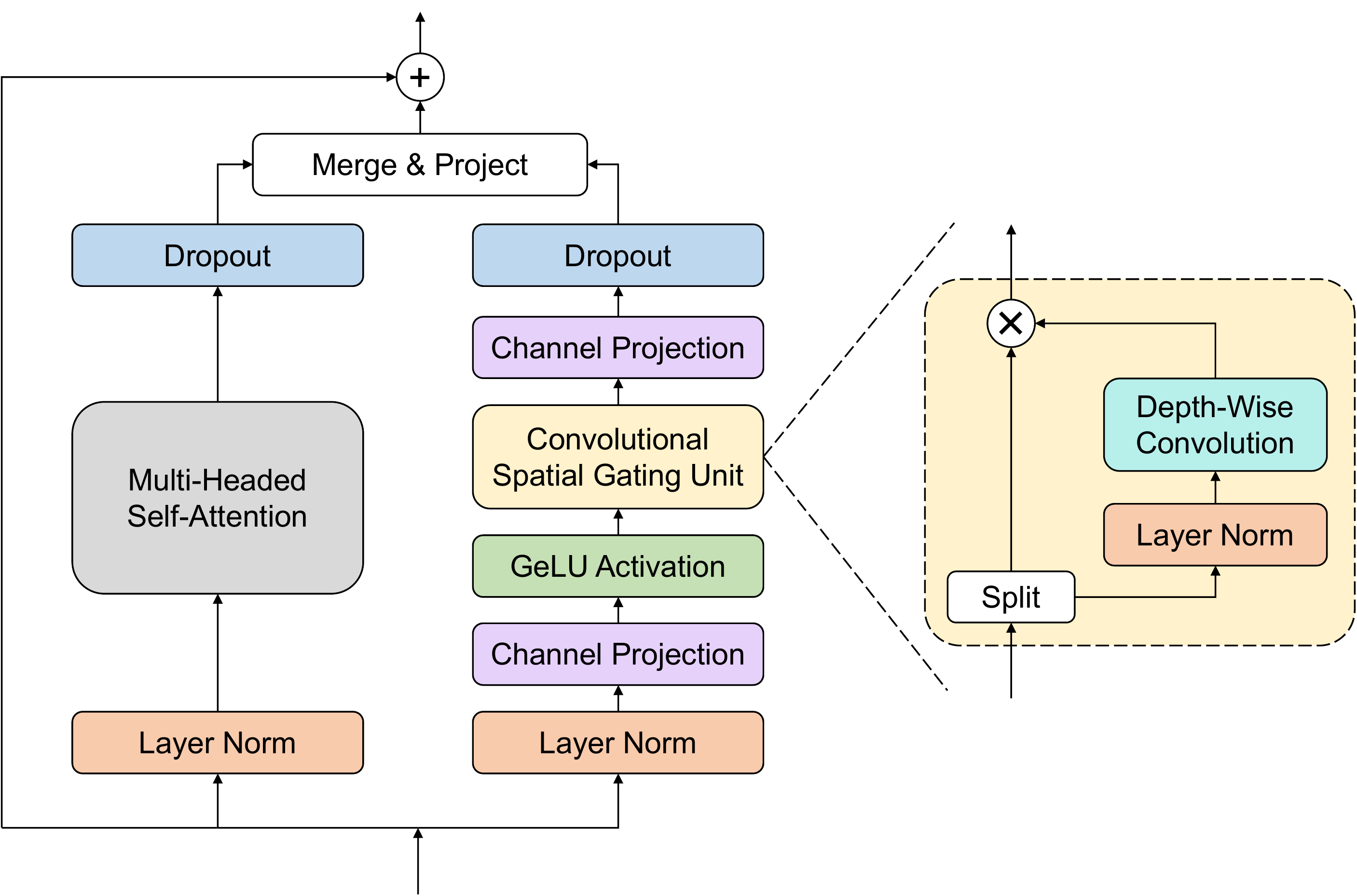}
\caption{Architecture of our Branchformer encoder block. It consists of two parallel branches. One branch employs attention to capture global context, while the other branch utilizes the MLP with convolutional gating to extract local context.}
\label{block-arch}
\end{subfigure}
\hfill
\caption{Architecture of the proposed Branchformer encoder and its components.}
\label{fig:model-arch}
\vskip -0.1in
\end{figure*}

\section{Introduction}

Conformer \cite{google-conformer}, which is a variant of Transformer \cite{transformer}, has been successfully applied to various speech processing tasks \cite{espnet-conformer, espnet-slu, conformer-speech-separation}. The key feature of Conformer is that it captures both local and global contextual information using convolution and self-attention, respectively. The effectiveness of combining local and global dependencies is also demonstrated by other studies in vision \cite{cvt} and language \cite{lite-transformer} tasks. Despite the superior performance of Conformer, there are some limitations. Conformer combines convolution and self-attention sequentially. This static single-branch architecture is hard to interpret and modify. It is unclear how local and global relationships are used in different encoder layers. Are they equally important in every layer? If not, which type of operation plays a major role in the initial layers? Understanding these questions can help researchers design better architectures.

To address these questions, we propose a flexible, interpretable and customizable encoder alternative, \textit{Branchformer}, with two parallel branches to capture various ranged context in end-to-end automatic speech recognition (ASR) and spoken language understanding (SLU) tasks. 
In Branchformer, one branch employs self-attention or its variant to capture long-range dependencies, while the other branch utilizes an advanced multi-layer perceptron (MLP) to capture local dependencies.
We adopt an MLP with gating called gMLP \cite{gMLP}, which is originally proposed in vision and language tasks and has been successfully applied to CTC-based speech recognition (cgMLP) \cite{iclr-submission-cgMLP}.

We conduct extensive experiments on three ASR and two SLU datasets. Results show that our model outperforms Transformer and cgMLP. It also matches with or outperforms the state-of-the-art Conformer. Furthermore, our Branchformer can be easily modified thanks to the disentangled two-branch design. The self-attention can be replaced by an efficient variant to reduce complexity with only minor performance degradation. If weighted average is used for branch merging, the learned weights represent the importance of each branch in different layers. This helps us understand how local and global relationships are utilized in a deep encoder model, which is beneficial for designing better architectures. 
Finally, by performing branch dropout during training, we can prune the model by removing the attention branch for faster inference speed without the need of re-training or fine-tuning of the Branchformer model. This allows practitioners to have two inference speeds within a single model and they can use either option depending on their need.

\section{Related Work}

\subsection{RNNs, CNNs, Transformers and MLPs}
\label{sec:relate-work-rnn-cnn-trans-mlp}
Recurrent neural networks (RNNs) have been widely used in speech processing \cite{listen-attend-spell, zeyer2018improved, specaugment}. RNNs are capable of modeling temporal dependencies in an audio sequence, but they are difficult to parallelize and the long-range interactions are not well captured. Convolutional neural networks (CNNs) are another key building block in modern deep learning. CNNs focus on local dependencies and are shift-invariant, which are especially suitable for vision tasks. They can also be applied to speech recognition \cite{cnn-lvcsr, jasper, contextnet}.

In the past few years, Transformers with self-attention \cite{transformer} have achieved superior performance in various speech applications \cite{transformer-vs-rnn-speech}. The self-attention mechanism is highly effective for modeling long-range global context, which is important for speech recognition and understanding. However, the time and memory complexity of self-attention is quadratic with respect to the sequence length. Researchers have proposed more efficient variants of self-attention \cite{efficient-transformers} such as Longformer \cite{longformer}, Big Bird \cite{bigbird}, Linformer \cite{linformer} and Fastformer \cite{fastformer}. These modifications are orthogonal to our work, so any of them can be utilized in our framework. We will evaluate the Fastformer model as an example, because it shows strong performance with linear complexity.

More recently, multi-layer perceptrons (MLPs) are revisited in language and vision tasks \cite{mlp-mixer, gMLP}. These MLP-based models have achieved comparable performance with Transformers, which is promising for further exploration. However, MLP-based models only accept fixed-length inputs, and thus cannot be directly applied to speech processing. To solve this problem, some MLP modules can be replaced by convolutions or other similar operations. For example, in \cite{iclr-submission-cgMLP}, three variants of MLP-based models are proposed for CTC-based speech recognition. Among them, the MLP with convolutional gating (abbreviated as cgMLP in this paper) achieves the best performance. In this work, we employ cgMLP to extract local dependencies.

In addition to these standard modules, prior studies have also explored the combination of different types of modules \cite{cldnn-tara, chen-combine-models, hao-combine-rnn-transformer}. Because different modules have complementary capacities, the combined model can potentially achieve better performance. Our Branchformer combines self-attention, convolution and MLP modules in a unified model.

\subsection{Modeling Both Local and Global Context}
Both local and global relationships play an important role in sequence modeling. Researchers have explored various methods to combine these two types of context in a unified model. Two well-known architectures are: Conformer \cite{google-conformer} and Lite Transformer \cite{lite-transformer}.

Conformer is a variant of Transformer, which consists of two Macaron-like feed-forward layers, a multi-headed self-attention module and a convolution module. These modules are combined sequentially, resulting in a static single-branch architecture. Conformer-based methods have achieved state-of-the-art performance in many speech applications \cite{google-conformer, espnet-conformer}. However, due to the single-branch design, it is difficult to analyze how local and global interactions are utilized in different layers. Conformer also enforces a fixed, interleaving pattern between self-attention and convolution, which may not always be optimal. As discovered in \cite{reordering-transformer}, reordering the self-attention and feed-forward layers can improve Transformer's performance. Another limitation of Conformer is the quadratic complexity with respect to the sequence length due to self-attention.

Similar to this work, Lite Transformer \cite{lite-transformer} also adopts a two-branch architecture based on the standard self-attention and convolution to capture global and local dependencies. However, the motivation is quite different from ours.\footnote{The motivation plays a big role in the architecture choices. For example, Lite Transformer splits the input along the feature dimension before feeding them separately into the two branches. This reduces computation, but can harm the capacity of both branches. For better mobile efficiency, Lite Transformer also adopts a ``flattened'' feed-forward network (FFN), which we do not need. Actually, our Branchformer does not have an explicit FFN after the two branches. Instead, the two linear layers (i.e., channel projections) are integrated into the cgMLP branch. This design is novel compared to most of prior models.}
Lite Transformer uses specialized parallel branches to reduce the overall computation and model size for mobile NLP applications.
In speech, convolutions play an integral role in its modeling due to the local correlations in continuous speech data, which Conformer has exploited.
Our Branchformer aims to re-design the architecture to make it more stable to train (\cref{sec-model-scaling-stability}), flexible to allow various attentions (\cref{efficient-attn-results}), interpretable to present interesting design analysis (\cref{analysis-branch-weights}), and have different inference complexity in a single model (\cref{model-pruning}).

\section{Branchformer}
\subsection{Overall Architecture with Two Parallel Branches}

The overall architecture of our Branchformer encoder is shown in \cref{encoder-arch}. The raw audio sequence is first processed by a \textbf{frontend} module to extract log Mel features. Then, a \textbf{convolutional subsampling} module is applied to downsample the feature sequence in time. Positional encodings are added to the subsampled feature sequence, which is useful for self-attention modules in the following \textbf{Branchformer encoder blocks}. We employ a stack of $N$ identical Branchformer blocks to capture both global and local relationships in the feature sequence.

The detailed architecture of the Branchformer encoder block is presented in \cref{block-arch}.
It consists of two parallel branches and a residual connection. The two branches share the same input, but focus on relationships of different ranges, which can be complementary to each other. In each branch, we first normalize the input using layer norm \cite{layer-norm}, then extract global or local dependencies using attention or cgMLP, respectively, which is followed by dropout \cite{dropout}. The outputs of two branches can be merged using concatenation or weighted average, with the original input added as a residual connection.

Compared with purely CNN or Transformer-based approaches, our Branchformer is able to capture both global and local context explicitly, which has been shown to be very important for various sequence processing tasks \cite{google-conformer, lite-transformer}, including ASR and SLU. Compared with the widely used Conformer which enforces a fixed interleaving pattern of self-attention and convolution operations, our two-branch architecture is more flexible, interpretable and customizable. As shown in our experiments (\cref{analysis-branch-weights}), the global and local contexts are not equally important in different encoder layers. Thus, a more flexible architecture may be more effective. We further evaluate strategies to reduce complexity, as discussed in \cref{efficient-attn-results} and \cref{model-pruning}.

\subsection{Attention Branch for Global Context Modeling}
\subsubsection{Multi-Headed Self-Attention}
\label{sec:self-attn}
In \cref{block-arch}, the left branch in Branchformer aims to model global context in the input sequence. We employ the multi-headed self-attention mechanism \cite{transformer} with relative positional encoding \cite{transformer-xl}. In self-attention, the input is $\mathbf{X}\in \mathbb{R}^{T\times d}$, where $T$ is the sequence length and $d$ is the feature size. The input matrix can be further transformed into query, key and value matrices $\mathbf{Q}, \mathbf{K}, \mathbf{V} \in \mathbb{R}^{T\times d}$. The scaled dot product is calculated between every query and key, which is usually formulated as matrix multiplication $\mathbf{Q}\mathbf{K}^T / \sqrt{d}$. This computation has quadratic complexity with respect to $T$, which is not suitable for long sequences. Then, the raw scores are normalized using softmax and the output of self-attention is a weighted combination of the values, which can be written as
\begin{align}
    \mathrm{Attention}(\mathbf{Q}, \mathbf{K}, \mathbf{V}) = \mathrm{softmax}\left(\frac{\mathbf{Q}\mathbf{K}^T}{\sqrt{d}}\right) \mathbf{V}.
\end{align}
In multi-headed self-attention, the queries, keys and values are projected $h$ times with different learnable linear layers. After that, the self-attention operation is performed in parallel for each of these projected versions. Their outputs are concatenated and transformed to the original size, which generates the final output, as shown below:
\begin{align*}
    \mathrm{MultiHead}(\mathbf{Q},\mathbf{K},\mathbf{V}) = \mathrm{concat}(\mathrm{head}_1\ldots\mathrm{head}_h)\mathbf{W}^O,
\end{align*}
where $\mathrm{head}_i = \mathrm{Attention}(\mathbf{Q}\mathbf{W}_i^Q, \mathbf{K}\mathbf{W}_i^K, \mathbf{V}\mathbf{W}_i^V)$, and the linear transforms are $\mathbf{W}_i^Q, \mathbf{W}_i^K, \mathbf{W}_i^V \in \mathbb{R}^{d\times d/h}$ and $\mathbf{W}^O \in \mathbb{R}^{d\times d}$.

\subsubsection{More Efficient Attention}
\label{sec:fastformer}
As discussed in \cref{sec:relate-work-rnn-cnn-trans-mlp}, many efficient variants of self-attention can be used to replace the standard self-attention in our two-branch architecture. Here, we evaluate the Fastformer-based \cite{fastformer} approach, which shows strong performance with linear complexity. The details of Fastformer are explained in \cref{sec:fastformer-arch} (\cref{fastformer-arch}). The key component is the \textbf{attention-based pooling}, which summarizes a sequence into a single vector with global context. Specifically, let $\mathbf{q}_1, \mathbf{q}_2, \ldots, \mathbf{q}_T \in \mathbb{R}^d$ be the input sequence of the attention pooling module, the output $\mathbf{q} \in \mathbb{R}^d$ is a weighted sum: $\mathbf{q} = \sum_{i=1}^T \alpha_i \mathbf{q}_i$, where $\alpha_i$ is the attention weight. To obtain these weights, we first calculate the scaled dot product between a learnable parameter vector $\mathbf{w} \in \mathbb{R}^d$ and every input $\mathbf{q}_i$, and then perform softmax normalization: 
\begin{align}
    \alpha_i = \frac{\exp(\mathbf{w}^T\mathbf{q}_i/\sqrt{d})}{\sum_{j=1}^T \exp(\mathbf{w}^T\mathbf{q}_j/\sqrt{d})}.
    \label{eq:attn-pooling-weights}
\end{align}

The complexity of attention-based pooling is linear with respect to the sequence length $T$, which is more efficient than self-attention and it is also capable of capturing global relationships.

\subsection{MLP Branch for Local Context Modeling}
\label{sec:mlp-branch}
In \cref{block-arch}, the right branch of Branchformer focuses on more localized context in a sequence. This is realized by the MLP with convolutional gating (cgMLP) module \cite{iclr-submission-cgMLP}, which employs depth-wise convolution and linear gating to learn strong representations from a sequence. cgMLP is more powerful than the standard convolution module in Conformer, which can potentially improve the capacity of our model.
The architecture of cgMLP is depicted in \cref{block-arch}. For brevity, the initial layer norm and the final dropout modules are omitted in the following formulation. The cgMLP module consists of a channel projection, an activation function such as Gaussian error Linear Unit (GeLU) \cite{GeLU}, a \textbf{convolutional spatial gating unit} (CSGU) and another channel projection. Specifically, given an input sequence $\mathbf{X}\in \mathbb{R}^{T\times d}$, the output is calculated as follows:
\begin{align}
    \mathbf{Z} &= \mathrm{GeLU}(\mathbf{X}\mathbf{U}) \in \mathbb{R}^{T\times d_{\mathrm{hidden}}},\\
    \tilde{\mathbf{Z}} &= \mathrm{CSGU}(\mathbf{Z}) \in \mathbb{R}^{T\times d_{\mathrm{hidden}}/2},\label{eq:csgu} \\
    \mathbf{Y} &= \tilde{\mathbf{Z}} \mathbf{V} \in \mathbb{R}^{T\times d},
\end{align}
where $\mathbf{U} \in \mathbb{R}^{d\times d_{\mathrm{hidden}}}, \mathbf{V} \in \mathbb{R}^{d_{\mathrm{hidden}}/2 \times d}$ denote the two channel projections. Usually, the hidden dimension $d_{\mathrm{hidden}}$ is larger than the original dimension $d$ (e.g., $d=256, d_{\mathrm{hidden}}=2048$), which is similar to the position-wise feed-forward layers in Transformer and Conformer.

The key component in cgMLP is the CSGU based on linear gating (\cref{eq:csgu}), which exploits a depth-wise convolution to capture local dependencies. First, the input feature sequence $\mathbf{Z} \in \mathbb{R}^{T\times d_{\mathrm{hidden}}}$ is equally split along the feature dimension, resulting in two new sequences $\mathbf{Z}_1, \mathbf{Z}_2 \in \mathbb{R}^{T\times d_{\mathrm{hidden}}/2}$. Then, $\mathbf{Z}_2$ is normalized with layer norm and processed by a depth-wise convolution along the time dimension: 
\begin{align}
    \mathbf{Z}_2^\prime = \mathrm{DWConv}(\mathrm{LayerNorm}(\mathbf{Z}_2)).
\end{align}
The final output of CSGU is the element-wise product of $\mathbf{Z}_1$ and $\mathbf{Z}_2^\prime$, i.e., $\tilde{\mathbf{Z}} = \mathbf{Z}_1 \otimes \mathbf{Z}_2^\prime$, where $\otimes$ denotes element-wise multiplication. This is a type of linear gating, as we do not use other nonlinear activations before the multiplication.

The computational costs of the two channel projections are $O(Tdd_{\mathrm{hidden}})$ and $O(Tdd_{\mathrm{hidden}}/2)$, respectively. The depth-wise convolution has $O(TKd_{\mathrm{hidden}}/2)$ complexity, where $K$ is the kernel size. The overall complexity is linear with respect to $T$, although there is a constant factor $K$.

\subsection{Merging Two Branches}
\label{sec:merging-operations}
We employ concatenation or weighted average to merge two branches. Concatenation is used as the default method since it is simple and effective, but weighted average is more interpretable. The branch weights indicate how global and local relationships are utilized in different layers.

\subsubsection{Concatenation}
\label{sec:concat-merging}
Let $\mathbf{Y}_{\mathrm{att}}, \mathbf{Y}_{\mathrm{mlp}} \in \mathbb{R}^{T\times d}$ be the output sequences from the attention branch and cgMLP branch, respectively. We first concatenate them along the feature dimension and then project the result back to the original dimension:
\begin{align}
    \mathbf{Y}_{\mathrm{merged}} = \mathrm{concat}(\mathbf{Y}_{\mathrm{att}}, \mathbf{Y}_{\mathrm{mlp}}) \mathbf{W}_{\mathrm{merge}} \in \mathbb{R}^{T\times d},
\end{align}
where $\mathbf{W}_{\mathrm{merge}} \in \mathbb{R}^{2d\times d}$ is a learnable matrix.

\subsubsection{Weighted Average}
\label{sec:weighted-ave}
The concatenation-based merging is very effective, but it is difficult to interpret and modify. To mitigate this issue, weighted average is also utilized in our work, where the weights are dynamically generated by the model. More specifically, we first summarize the output sequence of each branch into a single vector using the attention-based pooling as described in \cref{sec:fastformer}:
\begin{align}
    \mathbf{y}_{\mathrm{att}} &= \mathrm{AttPooling}(\mathbf{Y}_{\mathrm{att}}),\\
    \mathbf{y}_{\mathrm{mlp}} &= \mathrm{AttPooling}(\mathbf{Y}_{\mathrm{mlp}}),
\end{align}
where $\mathbf{y}_{\mathrm{att}}, \mathbf{y}_{\mathrm{mlp}} \in \mathbb{R}^{d}$. Then, the two vectors are projected to two scalars and normalized using softmax to get branch weights:
\begin{align}
    w_{\mathrm{att}}, w_{\mathrm{mlp}} = \mathrm{softmax}(\mathbf{W}_{\mathrm{att}} \mathbf{y}_{\mathrm{att}}, \mathbf{W}_{\mathrm{mlp}} \mathbf{y}_{\mathrm{mlp}}),
\end{align}
where $\mathbf{W}_{\mathrm{att}}, \mathbf{W}_{\mathrm{mlp}} \in \mathbb{R}^{1\times d}$ represent linear transforms. Finally, the merged output is a weighted average: $\mathbf{Y}_{\mathrm{merged}}^\prime = w_{\mathrm{att}} \mathbf{Y}_{\mathrm{att}} + w_{\mathrm{mlp}} \mathbf{Y}_{\mathrm{mlp}}$, where $\mathbf{Y}_{\mathrm{merged}}^\prime \in \mathbb{R}^{T\times d}$ captures both global and local dependencies.

To make it possible to prune the two-branch model for faster inference, we can drop the entire attention branch (i.e., setting the weight of the attention branch to zero and directly using the other branch) with a certain probability during training. We call this technique \textbf{branch dropout}, which is evaluated in \cref{model-pruning}.

\subsection{Complexity Analysis}
In this section, we analyze the complexity of the components in our proposed Branchformer. Let $T$ be the sequence length and $d$ be the feature dimension.

For the attention-based branch, we consider self-attention and Fastformer. In the standard self-attention formulation, the scaled dot product is calculated between all pairs of feature vectors in the sequence, which leads to $O(T^2d)$ complexity. In Fastformer, the attention-based pooling has compleixty $O(Td)$, as discussed in \cref{sec:fastformer}. The linear projections in both self-attention and Fastformer have complexity $O(Td^2)$. Fastformer is more efficient for longer sequences because its complexity is linear instead of quadratic w.r.t. the sequence length.

For the MLP-based branch, the overall computational cost is linear w.r.t. the sequence length, although there is a constant factor related to the kernel size, as discussed in \cref{sec:mlp-branch}.

\section{Experiments}
\subsection{Experimental Setup}
\subsubsection{Datasets}
We evaluate our models on ASR and SLU tasks.\footnote{We also tested the efficacy of Branchformer on machine translation. Results are shown in \cref{appsec:mt-espnet}.}
For ASR, three widely-used public datasets are used: (1) Aishell \cite{aishell-corpus}, which consists of 170 hours of Mandarin speech data; (2) Switchboard (SWBD) 300h \cite{swbd-corpus}, which contains about 300 hours of English telephone conversations, and (3) LibriSpeech 960h \cite{librispeech-corpus}, which consists of about 960 hours of English read audiobooks. For SLU, the recently released SLURP corpus \cite{slurp-corpus} is used for intent classification and entity prediction. SLURP is an English SLU dataset which is substantially larger and linguistically more diverse than previous SLU datasets. The Speech Commands dataset \cite{google-speech-commands-corpus} is also employed. The vocabulary contains 35 words. Each utterance is around one second.

\begin{table}[t]
\vskip -0.1in
\caption{Results presenting the \% Character Error Rate (CER) of our proposed Branchformer model on the Aishell Mandarin ASR task. Previously published papers and our reproduction of the baselines using cgMLP, Transformer, Lite Transformer and Conformer models are shown for comparison. No LM rescoring is used unless specified.}  
\label{aishell-results}
\begin{center}
\resizebox {0.9\linewidth} {!} {
\begin{tabular}{l @{\hspace{2\tabcolsep}} c @{\hspace{2\tabcolsep}} cc}
\toprule
Method & Params (M) & dev ($\downarrow$) & test ($\downarrow$) \\
\midrule
\multicolumn{4}{l}{\textit{SpeechBrain} \cite{speechbrain}}\\
\quad Transformer & -     & 5.60 & 6.04 \\
\multicolumn{4}{l}{\textit{WeNet} \cite{wenet}}\\
\quad Transformer & -     & -    & 5.30 \\
\quad Conformer   & -     & -    & \textbf{4.61} \\
\multicolumn{4}{l}{\textit{ESPnet} \cite{espnet-toolkit}}\\
\quad Transformer (+ RNN LM) & 30.4 & 5.9\hphantom{0} & 6.4\hphantom{0} \\
\quad Conformer   & 46.2 & \textbf{4.5\hphantom{0}} & 4.9\hphantom{0} \\
\midrule
\midrule
\multicolumn{4}{l}{\textit{Our Baselines} (reproduced based on ESPnet)}\\
\quad cgMLP        & 44.3 & 4.61 & 5.15 \\
\quad Transformer  & 46.1 & 4.83 & 5.17 \\
\quad Lite Transformer & 51.0 & 4.70 & 5.06 \\
\quad Conformer    & 46.3 & 4.24 & 4.62 \\
\midrule
\multicolumn{4}{l}{\textit{Our Proposed Model}}\\
\quad Branchformer & 45.4 & \textbf{4.19} & \textbf{4.43} \\
\bottomrule
\end{tabular}
}
\end{center}
\vskip -0.1in
\end{table}

\begin{table}[t]
\vskip -0.1in
\caption{Results presenting the \% Word Error Rate (WER) of our proposed Branchformer model on the Switchboard ASR task. Previously published papers and our reproduction of the baselines using Transformer, cgMLP and Conformer models are shown for comparison. No LM rescoring is used. $^{\diamondsuit}$Implemented with LSTM.}
\label{swbd-results}
\begin{center}
\resizebox {\linewidth} {!} {
\begin{tabular}{l @{\hspace{2\tabcolsep}} c @{\hspace{2\tabcolsep}} c @{\hspace{2\tabcolsep}} c @{\hspace{2\tabcolsep}} c}
\toprule
Method & Params (M) & swb ($\downarrow$) & chm ($\downarrow$) & eval2000 ($\downarrow$) \\
\midrule
\citeauthor{tuske2020swbd} \yrcite{tuske2020swbd}~$^\diamondsuit$ & -     & 7.6 & \textbf{14.6} & - \\
\citeauthor{specaugment} \yrcite{specaugment}~$^\diamondsuit$     & -     & \textbf{7.2} & \textbf{14.6} & - \\
\multicolumn{5}{l}{\textit{ESPnet} \cite{espnet-toolkit}}\\
\quad Transformer  & -     & 9.0 & 18.1 & 13.6 \\
\quad Conformer    & -     & \textbf{7.2} & \textbf{14.6} & \textbf{10.9} \\
\midrule
\midrule
\multicolumn{5}{l}{\textit{Our Baselines} (reproduced based on ESPnet)}\\
\quad cgMLP        & 32.6 & 8.7 & 16.3 & 12.5\\
\quad Transformer  & 44.4 & 9.0 & 16.0 & 12.5\\
\quad Conformer    & 44.5 & \textbf{7.8} & 14.5 & 11.1\\
\midrule
\multicolumn{5}{l}{\textit{Our Proposed Model}}\\
\quad Branchformer & 43.7 & \textbf{7.8} & \textbf{14.1} & \textbf{10.9}\\
\bottomrule
\end{tabular}
}
\end{center}
\end{table}

\begin{table}[t]
\vskip -0.1in
\caption{Results presenting the \% WER of our proposed Branchformer model on the LibriSpeech ASR task. Previously published papers and our reproduction of the Conformer baseline are shown for comparison. No LM rescoring is used unless specified.}
\label{librispeech-results}
\begin{center}
\resizebox {\linewidth} {!} {
\begin{tabular}{l @{\hspace{1\tabcolsep}} c @{\hspace{1\tabcolsep}} c @{\hspace{1\tabcolsep}} c @{\hspace{1\tabcolsep}} c @{\hspace{1\tabcolsep}} c}
\toprule
Method & Params (M) & \multicolumn{2}{c}{dev ($\downarrow$)} & \multicolumn{2}{c}{test ($\downarrow$)}\\\cmidrule(r){3-4}\cmidrule(r){5-6}
&  & clean & other & clean & other \\
\midrule
LSTM \cite{specaugment}    & -      & -   & -   & 2.8 & 6.8 \\
ContextNet \cite{contextnet}   & 112.7  & 2.0 & 4.6 & \textbf{2.1} & 4.6\\
Conformer \cite{google-conformer}    & 118.8  & \textbf{1.9} & \textbf{4.4} & \textbf{2.1} & \textbf{4.3} \\
\multicolumn{6}{l}{\textit{SpeechBrain} \cite{speechbrain}}\\
\quad Transformer + Transformer LM  & -   & -   & -   & 2.5 & 5.9 \\
\multicolumn{6}{l}{\textit{WeNet} \cite{wenet}}\\
\quad Conformer    & -      & -   & -   & 2.7 & 6.5 \\
\multicolumn{6}{l}{\textit{ESPnet} \cite{espnet-toolkit}}\\
\quad Transformer + RNN LM  & \hphantom{0}99.4          & 2.3 & 5.9 & 2.5 & 6.2 \\
\quad Conformer    & 116.2 & 2.3 & 6.1 & 2.6 & 6.0 \\
\midrule
\midrule
\multicolumn{6}{l}{\textit{Our Baseline} (reproduced based on ESPnet)}\\
\quad Conformer    & 116.2 & \textbf{2.2} & 5.6 & 2.5 & \textbf{5.5} \\
\midrule
\multicolumn{6}{l}{\textit{Our Proposed Model}}\\
\quad Branchformer & 116.2 & \textbf{2.2} & \textbf{5.5} & \textbf{2.4} & \textbf{5.5} \\
\bottomrule
\end{tabular}
}
\end{center}
\vskip -0.1in
\end{table}

\begin{table}[tb]
\vskip -0.1in
\caption{Results presenting the accuracy and SLU-F1 of our proposed Branchformer model on the SLURP SLU benchmark. Previously published papers and our reproduction of baselines using cgMLP, Transformer and Conformer are shown for comparison.}
\label{slurp-results}
\begin{center}
\resizebox {\linewidth} {!} {
\begin{tabular}{l @{} c @{\hspace{0.2\tabcolsep}} c @{\hspace{0.3\tabcolsep}} c @{\hspace{0.3\tabcolsep}} c}
\toprule
Method & \multicolumn{2}{c}{Intent Classification} & \multicolumn{2}{c}{Entity Prediction}\\\cmidrule(r){2-3}\cmidrule(r){4-5}
& Params (M) & Acc. ($\uparrow$) & Params (M) & SLU-F1 ($\uparrow$) \\
\midrule
\textit{SLURP Paper} \cite{slurp-corpus} & -   & 0.783 & -      & 0.708 \\
\multicolumn{4}{l}{\textit{SpeechBrain} \cite{speechbrain}}\\
\quad LSTM + pretrained ASR encoder & -      & 0.751 & -      & 0.633 \\
\multicolumn{4}{l}{\textit{ESPnet} \cite{espnet-slu}}\\
\quad Conformer    & 109.3 & \textbf{0.863} & -      & \textbf{0.719} \\
\midrule
\midrule
\multicolumn{5}{l}{\textit{Our Baselines} (reproduced based on ESPnet)}\\
\quad cgMLP        & 77.0  & 0.859 & 77.0  & 0.734 \\
\quad Transformer  & 90.2  & 0.872 & 90.3  & 0.717 \\
\quad Conformer    & 109.3 & 0.877 & 109.4 & 0.769 \\
\midrule
\multicolumn{5}{l}{\textit{Our Proposed Model}}\\
\quad Branchformer & 110.1 & \textbf{0.881} & 95.6  & \textbf{0.777} \\
\bottomrule
\end{tabular}
}
\end{center}
\vskip -0.2in
\end{table}

\subsubsection{Implementation Details}
Our models are implemented using PyTorch \cite{pytorch-toolkit} and the ASR and SLU experiments are conducted using the ESPnet toolkit \cite{espnet-toolkit, espnet-slu, espnet-conformer}. We follow the corresponding recipes in ESPnet for data preparation, model training and evaluation. The 80-dim log Mel filterbank features are extracted. The window and hop lengths vary for different datasets. SpecAugment \cite{specaugment} and speed perturbation are performed for data augmentation. We apply two $3\times3$ convolutions with a stride of 2 to subsample the original speech feature sequence. For different tasks, we adjust the feature size $d$, hidden size $d_{\mathrm{hidden}}$, number of layers and attention heads $h$ in the Branchformer encoder to make it comparable with previous models, but we always employ a 6-layer Transformer decoder. Our experiments are conducted using 4 Tesla V100 GPUs with 32GB memory. The Adam optimizer \cite{adam} with weight decay 1e-6 is utilized. We also apply dropout \cite{dropout} with probability 0.1 and label smoothing \cite{label-smoothing} with weight 0.1 to mitigate overfitting. We adopt the Transformer learning rate scheduler \cite{transformer}. The CTC weight is set to 0.3 for joint training with the attention decoder \cite{joint-ctc-att}. The joint CTC-attention decoding \cite{joint-ctc-att-decoding} is also performed. We average the best 10 checkpoints based on validation performance for inference. The exponential moving average technique is applied to the Switchboard corpus, but it is not used in other datasets. 
\cref{impl-details} in Appendix \ref{sec:imple_details} summarizes the hyper-parameters in each dataset for reproducibility.

\subsection{Main Results}
\label{sec:main-results}

In this section, we compare the proposed Branchformer encoder with previous studies. Here, we adopt the standard self-attention (\cref{sec:self-attn}) and concatenation-based merging (\cref{sec:concat-merging}). For fair comparisons, we reproduced cgMLP, Transformer and Conformer baselines using ESPnet based on our own environments.

\cref{aishell-results} shows the results on Aishell without a language model (LM). Compared with cgMLP and Transformer, our Branchformer reduces the character error rate (CER) by 0.7\% absolutely on the test set. It also outperforms Conformer by 0.2\%, achieving the best performance among these methods. 

\cref{swbd-results} presents the word error rates (WERs) on Switchboard 300h without an LM. Again, our Branchformer outperforms cgMLP and Transformer baselines by a large margin. It matches with Conformer on the Switchboard (swb) subset, and outperforms it on the CallHome (chm) subset by 0.4\%. The performance of Branchformer is also comparable with previous best results without LM fusion.

\cref{librispeech-results} compares WERs on LibriSpeech 960h. The full results are shown in \cref{tab:full-librispeech-results} of \cref{app:full-results-librispeech}. Our Branchformer achieves 2.4/5.5 without an LM and 2.1/4.5 with an LM on the test clean/other sets. This is comparable with the Conformer baseline, and is better than results reported by other open-source toolkits. Note that our numbers are worse than the best reported Conformer-Transducer result \cite{google-conformer} achieved by Google, because our model is based on a Transformer decoder instead of RNN-Transducer and our code-base is different.

\begin{figure}[t]
\centering
\resizebox {0.8\linewidth} {!} {
\begin{tikzpicture}
	\begin{axis}[
		xlabel=Number of Parameters (Millions),
		ylabel=Entity Prediction SLU-F1 ($\uparrow$),
		xtick={20,30,...,110},
		xmin=25,
	    every axis plot/.append style={thick},
        legend cell align={left},
        legend columns=1,
        legend style={at={(1,0.65)}, anchor=north east,nodes={scale=0.9, transform shape}}
		]
	\addplot[color=red, mark=triangle, mark options={scale=1.5}] coordinates {
	    (34.92,  0.7596)
	    (61.97,  0.7764)
	    (95.64,  0.7767)
	}; 
	\addplot[color=orange, mark=diamond, dashed, mark options={scale=1.5, solid}] coordinates {
		(34.04,  0.7278)
	    (54.3,   0.7371)
	    (77.04,  0.7343)
	};
	\addplot[color=brown!60!black,mark=square, dashed, mark options={scale=1.5, solid}] coordinates {
		(43.46,  0.7594)
	    (73.61,  0.7571)
	    (109.39, 0.7688)
	};
	\addplot[color=blue,mark=x, dashed, mark options={scale=1.5, solid}] coordinates {
		(35.45,  0.7247)
	    (60.43,  0.7233)
	    (90.25,  0.7174)
	};
    \legend{\underline{Branchformer},cgMLP,Conformer,Transformer}
	\end{axis}
\end{tikzpicture}
}
  \caption{SLURP Entity Prediction SLU-F1 vs. Model Size. Our Branchformer outperforms the previously proposed Conformer, cgMLP and Transformer models at all scales for the SLURP benchmark Spoken Language Understanding task.}
  \label{slurp-entity-scaling}
\end{figure}
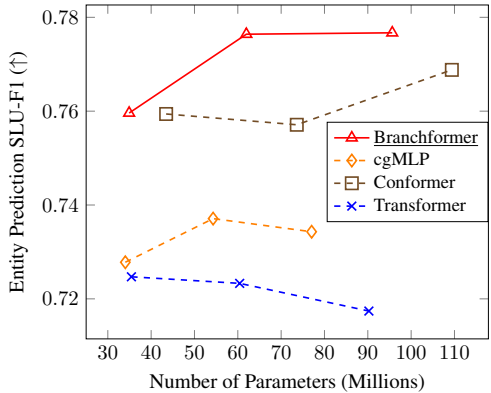

We evaluate our approach on two SLU tasks using SLURP, as shown in \cref{slurp-results}. Our Branchformer achieves the best accuracy for intent classification and the best SLU-F1 for entity prediction among the reported results, which demonstrates the effectiveness of our model in different tasks.

\subsection{Model Scalability and Training Stability}
\label{sec-model-scaling-stability}

\begin{table}[tb]
\vskip -0.1in
\caption{Accuracy performance of Branchformer vs. other architectures on Google Speech Commands (35 commands). Training the vanilla Conformer model is unstable on this dataset, but Branchformer achieves similar performance as other models.}
\label{speechcommands-results}
\vskip 0.1in
\begin{center}
\resizebox {\linewidth} {!} {
\begin{tabular}{l @{\hspace{\tabcolsep}} c @{\hspace{\tabcolsep}} c @{\hspace{1.5\tabcolsep}} c}
\toprule
Method & Params (M) & \multicolumn{2}{c}{Accuracy ($\uparrow$)} \\\cmidrule(r){3-4}
& & dev & test \\
\midrule
\multicolumn{4}{l}{\textit{SpeechBrain} \cite{speechbrain}}\\
\quad TDNN (+ xvector) & - & - & 0.974 \\
\multicolumn{4}{l}{\textit{ESPnet} \cite{espnet-slu}}\\
\quad Conformer (w/o BatchNorm) & - & \textbf{0.974} & \textbf{0.975} \\
\midrule
\midrule
\multicolumn{4}{l}{\textit{Our Baselines} (reproduced based on ESPnet)}\\
\quad cgMLP        & 30.7 & 0.966 & 0.966 \\
\quad Transformer  & 42.9 & \textbf{0.973} & \textbf{0.974} \\
\quad Conformer (w/ BatchNorm) & 43.0 & \multicolumn{2}{c}{diverged} \\
\midrule
\multicolumn{4}{l}{\textit{Our Proposed Model}}\\
\quad Branchformer & 41.8 & \textbf{0.973} & 0.973 \\
\bottomrule
\end{tabular}
}
\end{center}
\vskip -0.1in
\end{table}

We compare our Branchformer with baselines at different model scales. In \cref{aishell-scaling}, the CER on Aishell decreases for all models as the model size increases. At all three scales, the proposed Branchformer achieves the lowest CER. \cref{slurp-entity-scaling} shows the SLU-F1 scores on SLURP. Again, our model outperforms the baselines at all model scales. These results demonstrate the scalability and effectiveness of Branchformer. 

In our experiments, we have found that Branchformer is more stable to train than Conformer, especially on short utterances and limited data. This can be seen in our results on Google Speech Commands in \cref{speechcommands-results}. Our reproduction of the standard Conformer model diverged, which is consistent with the description reported in the ESPnet recipe \cite{espnet-slu}. This is probably due to the batch norm used in Conformer or the difficulty of optimizing deeper layers in Conformer (i.e., sequential combination of convolution and self-attention layers). Branchformer on the other hand is more stable and achieves similar results with other methods.

\subsection{Results of Efficient Attention}
\label{efficient-attn-results}

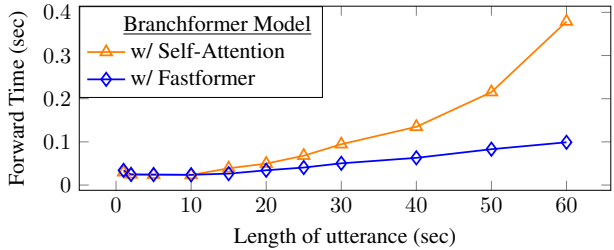
\begin{figure}[t]
    \centering
\resizebox {\linewidth} {!} {
    \begin{tikzpicture}
	\begin{axis}[
		xlabel=Length of utterance (sec),
		ylabel=Forward Time (sec),
		xtick={0,10,...,60},
	    every axis plot/.append style={thick},
        legend cell align={left},
        legend columns=1,
        legend style={at={(0,1)}, anchor=north west},
        height=4.5cm,
        width=10cm,
		]
	\addlegendimage{empty legend}\addlegendentry{\hspace{-.2cm} \underline{Branchformer Model}}
	\addplot[color=orange, mark=triangle, mark options={scale=1.5}] coordinates {
(1	,0.0297296283068135)
(2	,0.0246595375007018)
(5	,0.0240932676824741)
(10,	0.0233589647104963)
(15,	0.0388072846224531)
(20,	0.0496705639990977)
(25,	0.0679666062933392)
(30,	0.0945010547991842)
(40,	0.134878418687731)
(50,	0.215213371708523)
(60,	0.378710803284775)
	
	}; \addlegendentry{w/ Self-Attention}
	\addplot[color=blue, mark=diamond, mark options={scale=1.5}] coordinates {
(1	,0.034047583292704)
(2	,0.0244617013027891)
(5	,0.0242429405683651)
(10,	0.023771140014287)
(15,	0.0264533783774822)
(20,	0.0342037162743508)
(25,	0.0405612292932346)
(30,	0.0503172031138092)
(40,	0.0629996288800612)
(50,	0.0830339541891589)
(60,	0.0990663516218774)
	}; \addlegendentry{w/ Fastformer}
	\end{axis}
\end{tikzpicture}
}
\caption{Encoder forward time vs. input audio length using different attention mechanisms for modeling global dependencies in Branchformer. Branchformer w/ Fastformer achieves linear scaling in forward time with different utterance lengths.} 
\label{fig:time-fastformer-att}
\end{figure}

\begin{table}[tb]
\vskip -0.1in
\caption{Comparison of the Fastformer-based model with others on Aishell (\% CER) and Switchboard 300h (\% WER). Fastformer has linear complexity w.r.t. the sequence length $T$, while self-attention has quadratic complexity. $K$ denotes the convolution kernel size.}
\label{tab:efficient-results}
\centering
\vskip 0.1in
\resizebox {\linewidth} {!} {
\begin{tabular}{lccccc}
\toprule
Method & Complexity & \multicolumn{2}{c}{Aishell} & \multicolumn{2}{c}{SWBD 300h}\\\cmidrule(lr){3-4} \cmidrule(lr){5-6}
& & dev & test & swb & chm \\
\midrule
cgMLP       & $O(TK)$  & 4.61 & 5.15 & 8.7 & 16.3 \\ 
Transformer & $O(T^2)$ & 4.83 & 5.17 & 9.0 & 16.0 \\ 
Conformer   & $O(T^2)$ & 4.24 & 4.62 & \textbf{7.8} & 14.5 \\ 
\midrule
\multicolumn{6}{l}{Branchformer}\\
\quad w/ self-attention & $O(T^2)$ & \textbf{4.19} & \textbf{4.43} & \textbf{7.8} & \textbf{14.1} \\ 
\quad w/ Fastformer & $O(TK)$ & 4.22 & 4.58 & 7.9 & 14.5 \\ 
\bottomrule
\end{tabular}
}
\vskip -0.1in
\end{table}

As discussed in \cref{sec:fastformer}, the standard self-attention can be replaced by more efficient attention variants (e.g., Fastformer) to reduce complexity. We evaluate this approach on Aishell and Switchboard 300h. Results are presented in \cref{tab:efficient-results}. The encoder forward time for inputs of different lengths is shown \cref{fig:time-fastformer-att}. We randomly generate inputs with batch size 12 and run the encoder 10 times on a Tesla V100 GPU. Then the average time is reported. Branchformer with self-attention performs the best, but its complexity is quadratic in $T$. The Fastformer-based variant has linear complexity. Although its performance degrades slightly, it is still comparable with Conformer and is much better than cgMLP and Transformer. This demonstrates the flexibility of our architecture, thanks to the disentangled two-branch design and the strong cgMLP branch.\footnote{We also replaced the self-attention in Conformer with Fastformer and trained this efficient Conformer on Aishell. The CERs on dev and test sets are 4.69\% and 5.56\%, respectively, which are much worse than the original Conformer results.}

\subsection{Comparison of Merging Operations}

\begin{table}[tb]
\vskip -0.1in
\caption{Comparison of two methods for merging the branches of Branchformer on Aishell. Branchformer w/ concatenation performs better. Branchformer w/ weighted average slightly degrades performance but exhibits other desirable properties.} 
\label{tab:merging-results}
\vskip 0.1in
\begin{center}
\begin{small}
\begin{tabular}{lccc}
\toprule
Method & Params (M) & \multicolumn{2}{c}{CER (\%)}\\\cmidrule(r){3-4}
& & dev & test \\
\midrule
cgMLP        & 44.26 & 4.61 & 5.15 \\
Transformer  & 46.13 & 4.83 & 5.17 \\
Conformer    & 46.25 & 4.24 & 4.62 \\
\midrule
Branchformer \\
\quad w/ concatenation    & 45.43 & \textbf{4.19} & \textbf{4.43} \\
\quad w/ weighted average & 43.88 & 4.23 & 4.61 \\
\bottomrule
\end{tabular}
\end{small}
\end{center}
\vskip -0.2in
\end{table}

Two merging operations are discussed in \cref{sec:merging-operations}. \cref{tab:merging-results} compares their performance on Aishell. Weighted average is slightly worse than concatenation, but still matches with Conformer. In subsequent sections, we will show that weighted average is more intrepretable and customizable.

\subsection{Layer-Wise Analysis of Local/Global Branches}
\label{analysis-branch-weights}

\begin{figure}[t]
\begin{center}
\centerline{\includegraphics[width=\columnwidth]{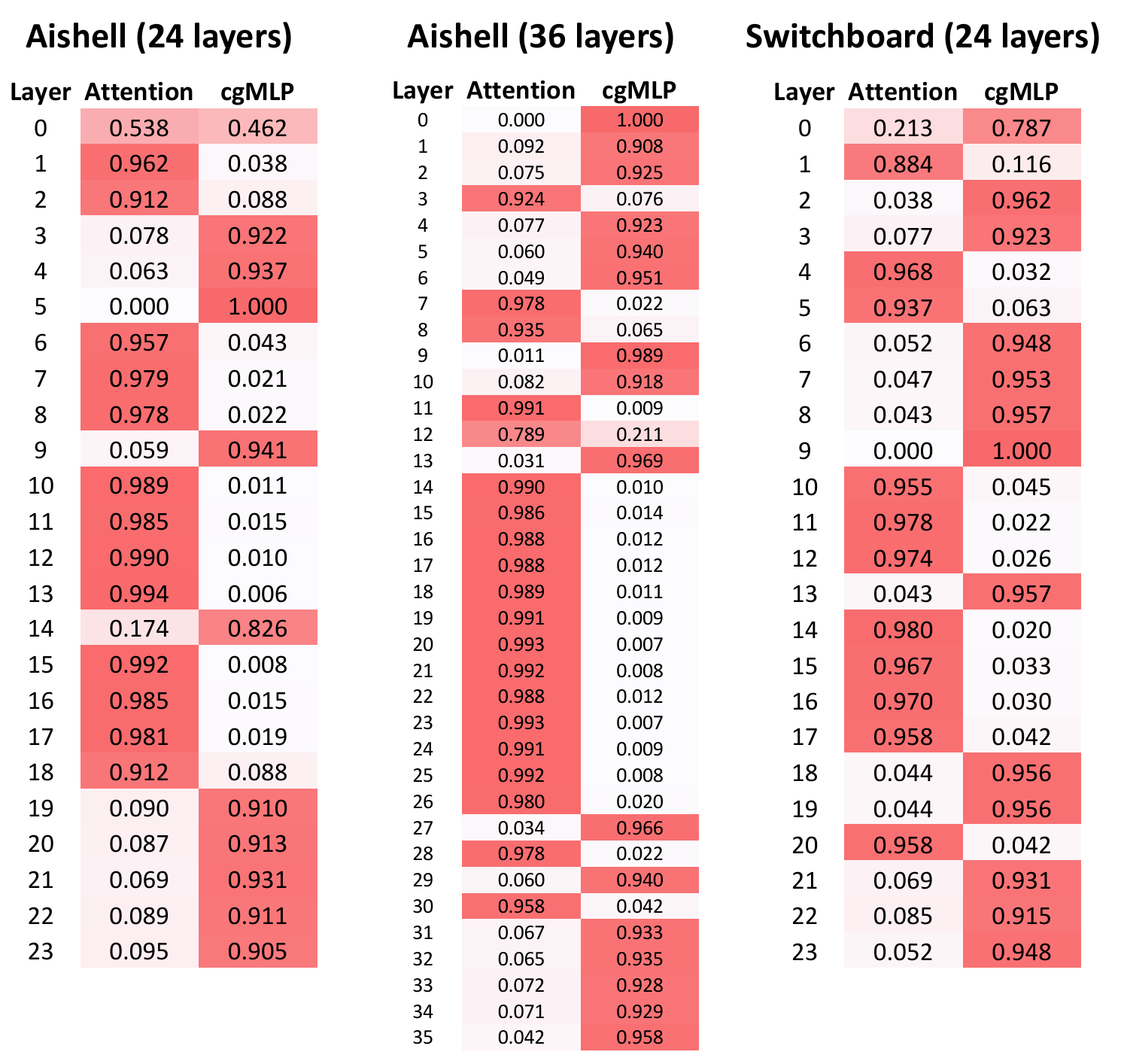}}
\caption{Visualization of branch weights in Branchformer with different sizes (24 vs. 36 layers) and on different datasets (Aishell vs. Switchboard). We see a consistent behavior of interleaving dominant branch in the beginning few layers followed by multiple consecutive layers of dominant global and local branches each.}
\label{fig:branch-weights}
\vskip -0.3in
\end{center}
\end{figure}

The learned weights for merging two branches introduced in Section \ref{sec:weighted-ave} can represent the importance of local and global context in different layers. 
We visualize the average weights on the validation set in \cref{fig:branch-weights}. The standard deviation is usually smaller than 0.01, which means the weights are very consistent for different samples. We can observe certain patterns in the weights. In almost all layers, one branch is dominant. In initial layers, the two types of branches are utilized in an interleaving fashion, which is similar to the Conformer design. This indicates the model is using both local and global relationships to learn strong hidden representations. In later layers, multiple consecutive attention blocks are observed, showing that global context is more important in the intermediate layers. Finally, multiple consecutive cgMLP blocks are leveraged to extract the local contextual information, which is then used by the Transformer decoder. Interestingly, similar patterns have also been discovered in prior studies. \citeauthor{cldnn-tara} have explored various ways to unify convolutional, recurrent and linear layers in a single model \yrcite{cldnn-tara} and found that it is better to employ RNNs for capturing global temporal dependencies in the middle stage and linear layers for capturing local interactions in the final stage. \citeauthor{reordering-transformer} has successfully improved Transformer by putting more self-attention blocks before feed-forward layers \yrcite{reordering-transformer}.

\begin{figure}[t]
\centering
\resizebox {\linewidth} {!} {
\begin{tikzpicture}
	\begin{axis}[
		xlabel=Layer,
		ylabel=Diagonality,
		xtick={0,1,...,23},
		xmin=0,
		xmax=23,
		ytick={0.5,0.6,...,1.0},
		ymin=0.5,
		ymax=1.0,
	    every axis plot/.append style={thick},
        legend cell align={left},
        legend columns=2,
        legend style={at={(1,1)}, anchor=north east},
        height=5cm,
        width=13cm,
		]
	\addplot[color=orange, mark=triangle, mark options={scale=1.5}] coordinates {
(0,     0.588)
(1,     0.785)
(2,     0.624)
(3,     0.594)
(4,     0.899)
(5,     0.884)
(6,     0.789)
(7,     0.636)
(8,     0.724)
(9,     0.752)
(10,    0.809)
(11,    0.729)
(12,    0.800)
(13,    0.830)
(14,    0.722)
(15,    0.730)
(16,    0.802)
(17,    0.812)
(18,    0.749)
(19,    0.628)
(20,    0.710)
(21,    0.721)
(22,    0.686)
(23,    0.596)
	}; \addlegendentry{Transformer}
	\addplot[color=blue, mark=diamond, mark options={scale=1.5}] coordinates {
(0 ,  0.618)
(1 ,  0.573)
(2 ,  0.556)
(3 ,  0.580)
(4 ,  0.612)
(5 ,  0.583)
(6 ,  0.585)
(7 ,  0.592)
(8 ,  0.524)
(9 ,  0.588)
(10,  0.584)
(11,  0.557)
(12,  0.558)
(13,  0.585)
(14,  0.588)
(15,  0.576)
(16,  0.551)
(17,  0.682)
(18,  0.582)
(19,  0.589)
(20,  0.607)
(21,  0.538)
(22,  0.564)
(23,  0.508)
	}; \addlegendentry{Branchformer}
	\end{axis}
\end{tikzpicture}
}
\vskip -0.1in
\caption{Diagonality of self-attention in each encoder layer. A higher diagonality means the attention weight matrix concentrates more on its diagonal, thus capturing more local context \cite{shucong-diagonality}.}
\label{fig:diagonality}
\vskip 0.1in
\end{figure}
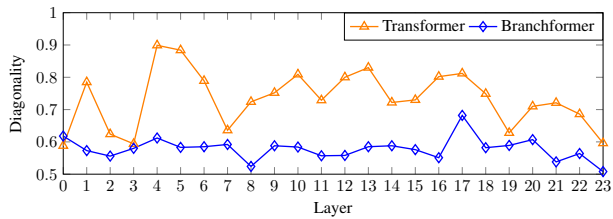

In the above discussion, we assume that the self-attention branch captures more global context in a sequence, but self-attention also has the potential to focus on local context. To analyze the behavior of self-attention in different architectures, we calculate the diagonality metric for the attention weight matrices in every encoder layer, as proposed by \citeauthor{shucong-diagonality} \yrcite{shucong-diagonality}. A higher diagonality means the attention weight matrix concentrates more on its diagonal, thus capturing more local context. We compute the average diagonality over all samples in the validation set and all attention heads in each layer. As shown in \cref{fig:diagonality}, Branchformer has lower diagonality than Transformer, showing that the self-attention branches of Branchformer capture more global relationships due to the disentangled two-branch design. Similar patterns of the self-attention weights have been observed in Lite Transformer \cite{lite-transformer}, because it also has specialized parallel branches. The definition of diagonality and some examples of the attention weights in Branchformer and Transformer can be found in \cref{appsec:vis-attn-weight-matrix}.

Based on the layer-wise analysis of two branches, the interleaving pattern of local and global blocks imposed by Conformer may not always be optimal. Our two-branch architecture dynamically determines the importance of each branch, which can be more flexible and powerful. To verify such observation and hypothesis, we combined the Conformer blocks with Branchformer blocks sequentially, resulting in a two-stage encoder model. Detailed results are presented in \cref{appsec:two-stage-models} (\cref{tab:two-stage-results}).
With Conformer blocks first and Branchformer blocks last, the performance is better than the vanilla Conformer and is similar to our Branchformer.

\begin{figure}[t]
\centering
\resizebox {\linewidth} {!} {
\begin{tikzpicture}
	\begin{axis}[
		xlabel=Length of utterance (sec),
		ylabel=Forward Time (sec),
		xtick={0,10,...,60},
	    every axis plot/.append style={thick},
        legend cell align={left},
        legend columns=1,
        legend style={at={(0,1)}, anchor=north west},
        height=4.5cm,
        width=10cm,
		]
	
	\addlegendimage{empty legend}\addlegendentry{\hspace{.2cm} \underline{Branchformer Model}}
	\addplot[color=orange, mark=triangle, mark options={scale=1.5}] coordinates {
(1	,0.0458660464151762)
(2	,0.0378558866796083)
(5	,0.0382390871993266)
(10,	0.0517894656863063)
(15,	0.0816142608178779)
(20,	0.114661365910433)
(25,	0.161982022202573)
(30,	0.212113050092011)
(40,	0.337697563704568)
(50,	0.502541744802147)
(60,	0.688907801371533)
	}; \addlegendentry{Original (both branches)}
	\addplot[color=blue, mark=diamond, mark options={scale=1.5}] coordinates {
(1	,0.0322461491916328)
(2	,0.0235273831058293)
(5	,0.0253130485070869)
(10,	0.0316343245911412)
(15,	0.0428912466973997)
(20,	0.0542070033145137)
(25,	0.0674742962000891)
(30,	0.0805344583699479)
(40,	0.103496388194616)
(50,	0.129218953428789)
(60,	0.153181256703101)
	}; \addlegendentry{Pruned (only cgMLP branch)}
	\end{axis}
\end{tikzpicture}
}
\vskip -0.1in
\caption{Encoder forward time vs. input audio length. The pruned model with a single branch has linear complexity, while the original two-branch model has quadratic complexity.}
\label{fig:model-pruning-time}
\end{figure}
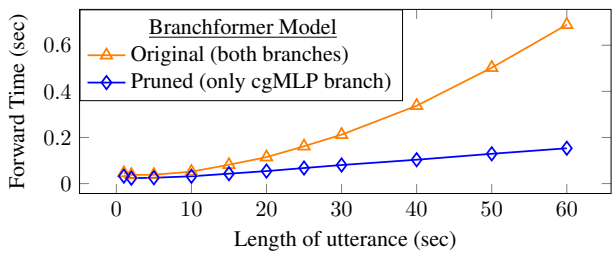

\subsection{Model Pruning Using Branch Dropout}
\label{model-pruning}

As described in \cref{sec:weighted-ave}, if we apply branch dropout to the attention branch during training, then our Branchformer can work in two modes for inference. In one mode, both branches are employed, which is more effective but slower. In the other mode, only the cgMLP branch is utilized, which reduces complexity from quadratic to linear and still keeps comparable performance with the same cgMLP trained from scratch. Note that this approach does not require fine-tuning or re-training. We simply train a single model with branch dropout, and it can work in two modes for inference. This demonstrates our two-branch model is more flexible and customizable. The encoder forward time for different input lengths is shown in \cref{fig:model-pruning-time}. We randomly generate an input with batch size 8 and run the encoder 10 times on a Tesla V100 GPU. The average time is reported. Detailed results with different dropout rates are shown in \cref{tab:branch-dropout-results} of \cref{appsec:branch-dropout}.

\section{Conclusion}

In this work, we present a novel encoder architecture called Branchformer. It has two parallel branches, one for capturing global interactions using attention and the other for more localized context using cgMLP. Due to the disentangled architecture, Branchformer can be customized easily by replacing the self-attention component with a more efficient attention. With the weighted average-based merging and branch dropout, we show that Branchformer can become flexible to have two different inference time complexity in a single trained model. By studying the branch weights in Branchformer, we could understand how local and global dependencies get utilized in different layers, which is helpful for model designing. Besides its flexible, interpretable, and customizable design, Branchformer outperforms Transformer and cgMLP by a large margin in various ASR and SLU benchmarks. It also achieves comparable or superior performance than the state-of-the-art Conformer while being more stable towards training in extreme data regimes.

\section*{Acknowledgements}
We thank the anonymous reviewers for their valuable feedback on our paper. This work used the Extreme Science and Engineering Discovery Environment (XSEDE) ~\cite{xsede}, which is supported by National Science Foundation grant number ACI-1548562. Specifically, it used the Bridges system ~\cite{nystrom2015bridges}, which is supported by NSF award number ACI-1445606, at the Pittsburgh Supercomputing Center (PSC).

\bibliography{example_paper}
\bibliographystyle{icml2022}

\newpage
\appendix
\onecolumn

\section{Fastformer Architecture}
\label{sec:fastformer-arch}

\begin{figure}[H]
\begin{center}
\centerline{\includegraphics[width=0.4\columnwidth]{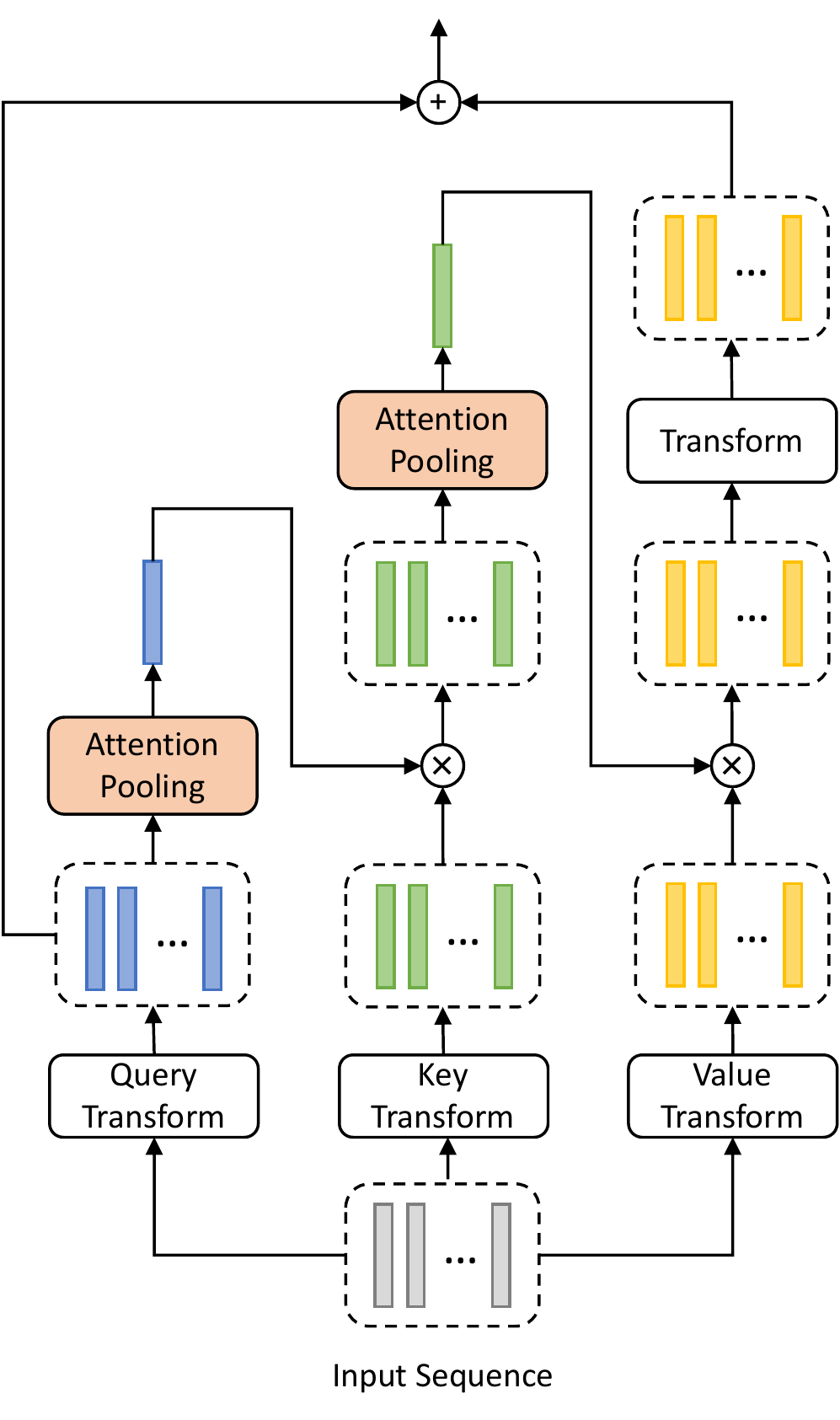}}
\caption{Architecture of the Fastformer block \cite{fastformer}, which is an efficient variant of Transformer with linear complexity. The attention-based pooling module extracts a global representation from a sequence based on attention mechanism, as shown in \cref{sec:fastformer} and \cref{eq:attn-pooling-weights}. $\otimes$ denotes element-wise multiplication.}
\label{fastformer-arch}
\end{center}
\vskip -0.2in
\end{figure}

\cref{fastformer-arch} shows the overall architecture of Fastformer \cite{fastformer}, which has linear complexity w.r.t. the sequence length. The attention-based pooling is described in \cref{sec:fastformer} and \cref{eq:attn-pooling-weights}.

In Fastformer, the input is first transformed into query, key and value sequences. Then, the attention pooling module extracts a single vector from the queries, which has global contextual information. The vector is multiplied with every key vector, resulting in a new sequence.

Next, we repeat the procedure described above for the new key sequence after multiplication. We perform another attention pooling to obtain a global representation for the new key sequence, and multiply it to every vector in the value sequence. This element-wise multiplication is followed by a final linear transform to generate the output of the entire Fastformer block. Note that we also add the query to the output, which is similar to a residual connection.

\section{Implementation Details}
\label{sec:imple_details}

The hyper-parameters used in our main experiments are presented in \cref{impl-details}.

\begin{table*}[htb]
\caption{Implementation details in different tasks and datasets. We show the Branchformer configurations used to generate main results in \cref{sec:main-results}. $T$ is the input sequence length.}
\label{impl-details}
\vskip 0.15in
\begin{center}
\begin{footnotesize}
\begin{tabular}{lcccccc}
\toprule
 & Aishell & Switchboard 300h & LibriSpeech 960h & \multicolumn{2}{c}{SLURP} & Speech Commands\\\cmidrule(lr){5-6}
 & & & & intent & entity \\
\midrule
\textit{Frontend}\\
\quad window length & 512 & 400 & 512 & 512 & 512 & 400\\
\quad hop length & 128 & 160 & 256 & 128 & 128 & 160\\
\midrule
\textit{SpecAug}\\
\quad time warp window & 5 & 5 & 5 & 5 & 5 & 5\\
\quad num of freq masks & 2 & 2 & 2 & 2 & 2 & 2\\
\quad freq mask width & (0, 27) & (0, 30) & (0, 27) & (0, 30) & (0, 30) & (0, 30)\\
\quad num of time masks & 10 & 2 & 10 & 2 & 2 & 2\\
\quad time mask width & (0, 0.05$T$) & (0, 40) & (0, 0.05$T$) & (0, 40) & (0, 40) & (0, 40)\\
\midrule
\textit{Architecture}\\
\quad feature size $d$ & 256 & 256 & 512 & 512 & 512 & 256\\
\quad hidden size $d_{\mathrm{hidden}}$ & 2048 & 2048 & 2048 & 3072 & 2048 & 2048\\
\quad attention heads $h$ & 4 & 4 & 8 & 8 & 8 & 4\\
\quad num of encoder layers & 24 & 24 & 22 & 18 & 18 & 24\\
\quad depth-wise conv kernel & 31 & 31 & 31 & 31 & 31 & 31\\
\midrule
\textit{Training}\\
\quad epochs & 60 & 90 & 60 & 50 & 50 & 150\\
\quad learning rate & 1e-3 & 2e-3 & 2.5e-3 & 1e-3 & 1e-3 & 3e-4\\
\quad warmup steps & 35k & 50k & 40k & 35k & 35k & 15k\\
\quad weight decay & 1e-6 & 1e-6 & 1e-6 & 1e-6 & 1e-6 & 1e-6\\
\quad dropout rate & 0.1 & 0.1 & 0.1 & 0.1 & 0.1 & 0.1\\
\quad ctc weight & 0.3 & 0.3 & 0.3 & 0.3 & 0.3 & 0\\
\quad label smoothing weight & 0.1 & 0.1 & 0.1 & 0.1 & 0.1 & 0.1\\
\quad exponential moving average & NA & 0.9999 & NA & NA & NA & NA\\
\bottomrule
\end{tabular}
\end{footnotesize}
\end{center}
\vskip -0.1in
\end{table*}

\section{Full Results on LibriSpeech 960h}
\label{app:full-results-librispeech}
\cref{tab:full-librispeech-results} shows the full results on LibriSpeech 960h, both with and without LMs. A brief version is in \cref{librispeech-results}.

\begin{table}[tb]
\caption{WERs (\%) on the LibriSpeech 960h dataset. Previous studies and our reproduced baselines are shown for comparison. The Transformer LM used in our experiments was downloaded from ESPnet.}
\label{tab:full-librispeech-results}
\vskip 0.15in
\begin{center}
\begin{tabular}{lcccccc}
\toprule
Method & Params (M) & LM & \multicolumn{4}{c}{WER (\%)}\\\cmidrule(r){4-7}
& & & \multicolumn{2}{c}{dev} & \multicolumn{2}{c}{test}\\\cmidrule(r){4-5}\cmidrule(r){6-7}
& & & clean & other & clean & other \\
\midrule
\multicolumn{7}{l}{\textit{\citeauthor{specaugment}} \yrcite{specaugment}}\\
\quad LSTM         & -      & -           & -   & -   & 2.8 & 6.8 \\
\quad LSTM         & -      & RNN         & -   & -   & 2.5 & 5.8 \\
\multicolumn{7}{l}{\textit{\citeauthor{contextnet}} \yrcite{contextnet}}\\
\quad ContextNet   & 112.7  & -           & 2.0 & 4.6 & 2.1 & 4.6\\
\quad ContextNet   & 112.7  & LSTM        & -   & -   & 1.9 & 4.1\\
\multicolumn{7}{l}{\textit{\citeauthor{google-conformer}} \yrcite{google-conformer}}\\
\quad Conformer    & 118.8  & -           & 1.9 & 4.4 & 2.1 & 4.3 \\
\quad Conformer    & 118.8  & LSTM        & -   & -   & 1.9 & 3.9 \\
\midrule
\multicolumn{7}{l}{\textit{SpeechBrain} \cite{speechbrain}}\\
\quad Transformer  & -      & Transformer & -   & -   & 2.5 & 5.9 \\
\multicolumn{7}{l}{\textit{WeNet} \cite{wenet}}\\
\quad Conformer    & -      & -           & -   & -   & 2.7 & 6.5 \\
\quad Conformer    & -      & 4-gram      & -   & -   & 2.7 & 6.0 \\
\multicolumn{7}{l}{\textit{ESPnet} \cite{espnet-toolkit}}\\
\quad Transformer  & 99.4  & RNN         & 2.3 & 5.9 & 2.5 & 6.2 \\
\quad Conformer    & 116.2 & -           & 2.3 & 6.1 & 2.6 & 6.0 \\
\quad Conformer    & 116.2 & Transformer & 1.9 & 4.6 & 2.1 & 4.7 \\
\midrule
\midrule
\multicolumn{7}{l}{\textit{Our Baselines} (reproduced based on ESPnet)}\\
\quad cgMLP & 108.0 & - & 2.3 & 6.3 & 2.6 & 6.1 \\
\quad Transformer & 109.6 & - & 2.5 & 6.5 & 2.8 & 6.5 \\
\quad Conformer & 116.2 & - & \textbf{2.2} & 5.6 & 2.5 & \textbf{5.5}\vspace{0.7em}\\
\quad cgMLP & 108.0 & Transformer & 1.9 & 4.5 & \textbf{2.1} & 4.6 \\
\quad Transformer & 109.6 & Transformer & 2.1 & 4.7 & 2.3 & 5.0 \\
\quad Conformer & 116.2 & Transformer & \textbf{1.8} & 4.3 & \textbf{2.1} & \textbf{4.5} \\
\midrule
\multicolumn{7}{l}{\textit{Our Proposed Model}}\\
\quad Branchformer & 116.2 & -           & \textbf{2.2} & \textbf{5.5} & \textbf{2.4} & \textbf{5.5}\vspace{0.7em}\\
\quad Branchformer & 116.2 & Transformer & 1.9 & \textbf{4.2} & \textbf{2.1} & \textbf{4.5} \\
\bottomrule
\end{tabular}
\end{center}
\vskip -0.1in
\end{table}

\section{Diagonality of Attention Weight Matrices}
\label{appsec:vis-attn-weight-matrix}

\cref{analysis-branch-weights} analyzes the diagonality metric \cite{shucong-diagonality} of the self-attention weight matrices in different encoder layers, which shows that the self-attention branches of Branchformer extract more global context compared with the original Transformer.

We follow the definition of diagonality in \cite{shucong-diagonality}. Let $\mathbf{A}\in\mathbb{R}^{T\times T}$ denote an attention weight matrix, where $T$ is the sequence length. Each element $a_{ij}$ is the attention weight between the $i$th and $j$th feature vectors in the sequence. Each row is a probability distribution, and the weights in each row sum to 1. Let's first define the centrality $C_i$ of the $i$th row:
\begin{align}
    C_i = 1 - \frac{\sum_{j=1}^T a_{ij}\vert i - j\vert}{\max\limits_{1\le j\le T} \vert i - j \vert}.
\end{align}
Then, the diagonality $D$ of the entire matrix is defined as the average centrality over all rows:
\begin{align}
    D = \frac{1}{T} \sum_{i=1}^T C_i.
\end{align}

\begin{figure*}[htb]
\centering
\begin{subfigure}[b]{\textwidth}
\centering
\includegraphics[width=0.66\textwidth]{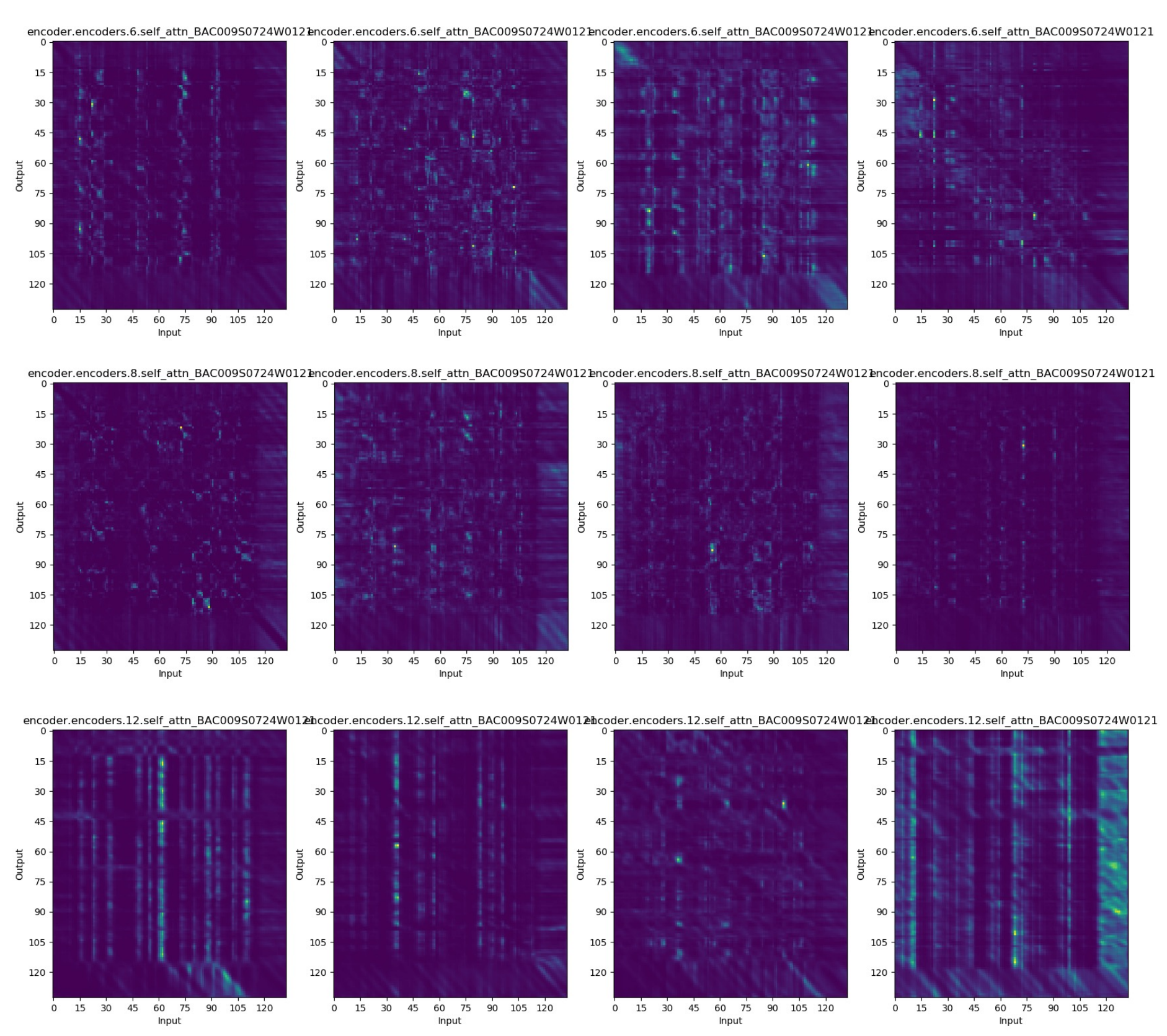}
\caption{Branchformer}
\label{appfig:attn-weight-branchformer}
\end{subfigure}
\begin{subfigure}[b]{\textwidth}
\centering
\includegraphics[width=0.65\textwidth]{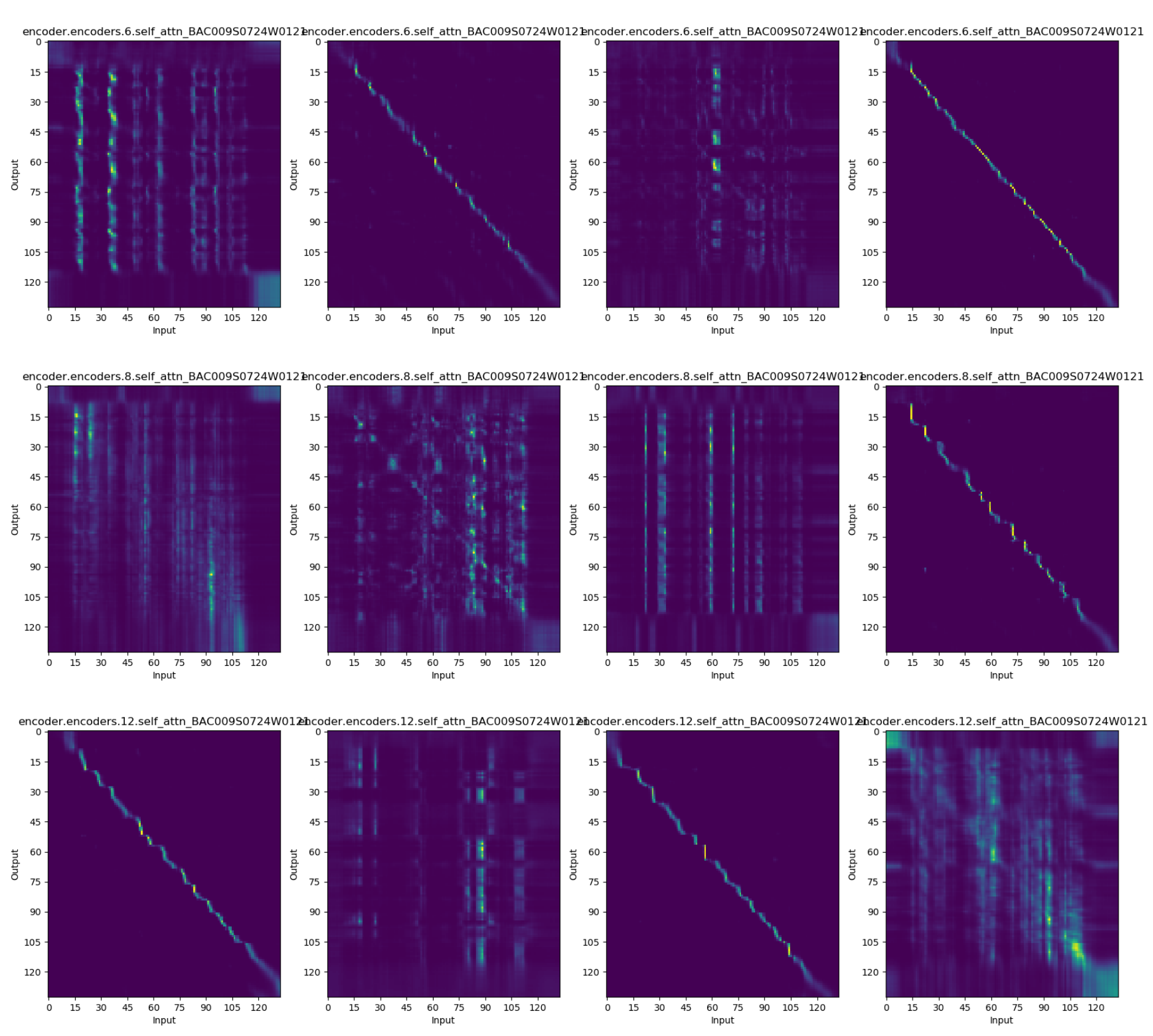}
\caption{Transformer}
\label{appfig:attn-weight-transformer}
\end{subfigure}
\caption{Examples of the self-attention weights in Branchformer and Transformer. Transformer has more diagonal attention weights, which means the attention blocks are capturing more local dependencies in a sequence.}
\label{appfig:attn-weights}
\end{figure*}

\cref{appfig:attn-weights} plots some attention weights from Branchformer and Transformer. We can see that Transformer has more diagonal attention weights, which means these attention heads are capturing local context. Our Branchformer has fewer diagonal patterns, showing that it extracts more global relationships.

\section{Results of Two-Stage Mixed Models}
\label{appsec:two-stage-models}

As mentioned in \cref{analysis-branch-weights}, we show the results using the two-stage mixed encoder. Specifically, we combined the Conformer blocks with Branchformer blocks sequentially, resulting in a two-stage encoder model. Results are presented in \cref{tab:two-stage-results}. 

If we use Conformer blocks first and Branchformer blocks later, the performance is better than the vanilla Conformer and is similar to our Branchformer. However, if the order is reversed, the performance degrades. These results verify our previous analysis. It is better to have interleaving local and global blocks in earlier layers and have more specialized blocks in later layers.

\begin{table}[tb]
\caption{Results of the two-stage encoder on Aishell.}
\label{tab:two-stage-results}
\vskip 0.15in
\begin{center}
\begin{tabular}{lcccc}
\toprule
Method & Layers & Params (M) & \multicolumn{2}{c}{CER (\%)}\\\cmidrule(lr){4-5}
& & & dev & test \\
\midrule
\multicolumn{5}{l}{\textit{Single-stage (using one type of encoder)}}\\
\quad Conformer    & 12 & 46.3 & 4.24 & 4.62 \\
\quad Branchformer & 24 & 45.4 & 4.19 & \textbf{4.43} \\
\midrule
\multicolumn{5}{l}{\textit{Two-stage (using two types of encoders sequentially)}} \\
\quad Conformer + Branchformer & 6+12 & 45.8 & \textbf{4.18} & 4.47 \\
\quad Branchformer + Conformer & 12+6 & 45.8 & 4.35 & 4.57\\
\bottomrule
\end{tabular}
\end{center}
\vskip -0.1in
\end{table}

\section{Effects of Branch Dropout}
\label{appsec:branch-dropout}

The branch dropout is introduced in \cref{sec:weighted-ave}. The results and discussions can be found in \cref{model-pruning}. We apply dropout to the attention branch during training so that we can prune the two-branch model for inference. This approach does not need fine-tuning or re-training. The trained model can operate in two different speeds, with one being linear and the other being quadratic w.r.t. the sequence length.

There is a trade-off between the performance of the two modes. As shown in \cref{tab:branch-dropout-results}, as we increase the dropout rate, the original model degrades in performance, but the pruned model improves. With a relatively large dropout rate, the pruned model is comparable with the same model trained from scratch.

The inference time is shown in \cref{fig:model-pruning-time} in \cref{model-pruning}.

\begin{table}[tb]
\caption{CERs (\%) on Aishell with different dropout rates for the self-attention branch. The two branches are merged using weighted average. We apply branch dropout during training. For inference, we can employ the original Branchformer model with both branches, or prune it by removing the attention branch.}
\label{tab:branch-dropout-results}
\vskip 0.15in
\begin{center}
\begin{tabular}{ccccc}
\toprule
Branch Dropout Rate & \multicolumn{2}{c}{Original Model} & \multicolumn{2}{c}{Pruned Model}\\\cmidrule(r){2-3} \cmidrule(r){4-5}
& dev & test & dev & test \\
\midrule
0.0 & 4.23 & 4.61 & 98.74 & 98.78 \\
\midrule
0.3 & 4.23 & 4.63 & 5.09 & 6.13 \\
0.5 & 4.31 & 4.72 & 4.85 & 5.63 \\
0.6 & 4.30 & 4.65 & 4.68 & 5.29 \\
0.7 & 4.40 & 4.80 & 4.67 & 5.22 \\
0.8 & 4.46 & 4.91 & 4.60 & 5.10 \\
\midrule
\midrule
\multicolumn{1}{l}{Conformer}  & 4.24 & 4.62 & - & - \\
\multicolumn{1}{l}{cgMLP}      & 4.61 & 5.15 & - & - \\
\bottomrule
\end{tabular}
\end{center}
\vskip -0.1in
\end{table}

\section{Preliminary Results on Machine Translation}
\label{appsec:mt-espnet}

Our Branchformer is a general encoder model which might be useful for other sequence modeling tasks in addition to speech processing. We have conducted preliminary experiments on the neural machine translation (NMT) task using ESPnet. The dataset is IWSLT 14 De-En, which has been widely used for evaluating NMT systems. The results are presented in \cref{tab:mt-preliminary-results}. Our proposed Branchformer achieves higher BLEU scores than the standard Transformer. We also observed that Branchformer converged faster during training. Note that the text input sequence is much shorter than a speech feature sequence, so we need to use a smaller convolution kernel size (e.g., around 5) in the cgMLP branch.

\begin{table}[htb]
\caption{BLEU scores of the machine translation task on a widely used corpus, IWSLT 14 De-En. Our experiments are based on ESPnet.}
\label{tab:mt-preliminary-results}
\vskip 0.15in
\begin{center}
\begin{tabular}{lccc}
\toprule
Method & Params (M) & \multicolumn{2}{c}{BLEU}\\
& & valid & test\\
\midrule
\multicolumn{4}{l}{\textit{\citeauthor{sid-mt}} \yrcite{sid-mt}} \\
\quad Transformer & 42 & 33.44 & 32.15 \\
\midrule
\multicolumn{4}{l}{\textit{Our Results Using ESPnet}}\\
\quad Transformer & 41.8 & 34.10 & 33.13 \\
\quad Branchformer & 43.4 & \textbf{34.96} & \textbf{33.72} \\
\bottomrule
\end{tabular}
\end{center}
\vskip -0.1in
\end{table}


\end{document}